\pdfoutput=1

\documentclass[11pt]{article}

\usepackage{emnlp2021}

\usepackage{times}
\usepackage{latexsym}

\usepackage[T1]{fontenc}

\usepackage[utf8]{inputenc}

\usepackage{microtype}
\usepackage{adjustbox}
\usepackage[font=small]{caption}
\usepackage{parskip}
\usepackage{booktabs}

\newcommand{\cut}[1]{}

\title{Exploring Strategies for Generalizable Commonsense Reasoning\\ with Pre-trained Models}

\author{
Kaixin Ma$^{\dagger}$,
Filip Ilievski$^{\S}$,
Jonathan Francis$^{\dagger\P}$,\\ 
\textbf{Satoru Ozaki$^{\dagger}$,
Eric Nyberg$^{\dagger}$,
Alessandro Oltramari$^{\P}$}\\
  $^{\dagger}$Language Technologies Institute, Carnegie Mellon University\\
  $^{\S}$Information Sciences Institute, University of Southern California\\
  $^{\P}$Human-Machine Collaboration, Bosch Research Pittsburgh\\
\small\{kaixinm, jmf1, sozaki, ehn\}@cs.cmu.edu, ilievski@isi.edu, alessandro.oltramari@us.bosch.com
\vspace{2mm}
}

\begin{document}
\maketitle
\begin{abstract}
Commonsense reasoning benchmarks have been largely solved by fine-tuning language models\cut{, i.e., updating its weights based on task-specific training data}. The downside is that fine-tuning may cause models to overfit to task-specific data and thereby forget their knowledge gained during pre-training. Recent works only propose lightweight model updates as models may already possess useful knowledge from past experience, but a challenge remains in understanding what parts and to what extent models should be refined for a given task. 
In this paper, we investigate what models learn from commonsense reasoning datasets. We measure the impact of three different adaptation methods on the generalization and accuracy of models. Our experiments with two models show that fine-tuning performs best, by learning both the content and the structure of the task, but suffers from overfitting and limited generalization to novel answers. We observe that alternative adaptation methods like prefix-tuning have comparable accuracy, but generalize better to unseen answers and are more robust to adversarial splits.

\end{abstract}


\section{Introduction}
\label{sec:intro}


Machine commonsense reasoning has recently gained new traction, largely due to a collection of diverse benchmarks~\cite{talmor-etal-2019-commonsenseqa, bhagavatula2019abductive, sap-etal-2019-social} and the successful application of language modeling methods on these benchmarks~\cite{ma-etal-2019-towards, shwartz-etal-2020-unsupervised, bauer-bansal-2021-identify}.
The most widely adopted approach to solve these commonsense reasoning tasks is by fine-tuning large pre-trained language models (LMs)~\cite{devlin-etal-2019-bert, Liu2019RoBERTaAR} on the task-specific training data. Meanwhile, it has been shown that language models are able to acquire certain commonsense background knowledge, during their pre-training on large textual data~\cite{petroni2019language, davison-etal-2019-commonsense, ma2020knowledgedriven}. In light of these findings and the large capacity of these language models, recent work has proposed lightweight alternatives to fine-tuning LMs, e.g., by only updating a small amount of additional parameters~\cite{lin-etal-2020-exploring, li2021prefixtuning}, or by updating the inputs while keeping the model weights intact~\cite{jiang-etal-2020-know, autoprompt:emnlp20}. Intuitively, these lightweight methods may retain the model's pre-trained knowledge to a large extent, and elicit the suitable knowledge for the target task, provided that much of this knowledge has already been encoded in the model parameters. However, to our knowledge, no comprehensive comparison exists between these model updating strategies. 

In this paper, we pose the question: \textit{What do models learn from commonsense reasoning datasets?} We consider three representative learning methods: regular fine-tuning, model extension with prefix-tuning \cite{li2021prefixtuning}, and model prompting with Autoprompt \cite{autoprompt:emnlp20}. We apply them to two representative model classes: the auto-regressive language model GPT-2 \cite{radford2019language} and sequence-to-sequence language model BART \cite{lewis-etal-2020-bart}. We conduct thorough evaluation on the generative evaluation benchmarks ProtoQA~\cite{boratko-etal-2020-protoqa} and CommonGen \cite{lin-etal-2020-commongen}, by training on different partitions of the training data. Our experiments show that fine-tuning performs best, by learning both the content and the structure of the task, but suffers from overfitting and limited generalization to novel answers. Prompting methods have lower accuracy, but tend to \cut{generalize better to unseen answers and to} show higher robustness to ``adversarial'' splits. Extending the models by prefix-tuning represents a ``sweet spot'' between task accuracy, generalization, and robustness. 

\section{Related Work}
Prior works probe the commonsense knowledge learned by the LMs. \citet{davison-etal-2019-commonsense} mined commonsense knowledge from LMs, using templates with masked tokens; \citet{richardson2020does} designed diagnostic tasks to probe LMs' knowledge of definitions and taxonomic reasoning. \cut{They observe that while LMs possess certain types of structured knowledge, their performance degrades with a slight increase in the number of inference steps or in the confounding strength of the distractors.} The LAMA probes~\cite{petroni2019language} demonstrate that LMs can largely recover knowledge in existing (commonsense) knowledge graphs: they could thus be queried/prompted directly as knowledge bases~\cite{shwartz-etal-2020-unsupervised,autoprompt:emnlp20}\cut{, or as knowledge models that can flexibly complete existing knowledge graphs~\cite{hwang2020comet,Bosselut2019DynamicKG}}. \citet{ettinger2020bert} diagnoses the BERT model, finding that \cut{it can generally distinguish good from bad completions and understand noun hypernymy, but}it struggles with complex inference, role-based event prediction, and grasping the contextual impacts of negation.\cut{The analysis of the aNLI benchmark~\cite{bhagavatula2019abductive} reveals that language models perform relatively well, off-the-shelf, but struggle with numeric and spatial inference.} The logical commonsense probes in RICA~\cite{zhou2020rica} show that LMs perform similar to random guessing in the zero-shot setting, they are heavily impacted by statistical biases, and are not robust to linguistic perturbations. \citet{elazar2021back} posit that while LMs can learn to perform well on commonsense tasks, their commonsense reasoning ability mostly comes from fine-tuning on the task data. \\
Some works have sought to uncover what models learn through training on question answering datasets, exposing various dataset artifacts in the process \cite{jia-liang-2017-adversarial, kaushik-lipton-2018-much,pugaliya-etal-2019-bend}. \citet{DBLP:journals/corr/abs-2003-04808} found that models trained on the SQuAD2.0 dataset \cite{rajpurkar-etal-2018-know} are insensitive to the meaningful changes in the question and predict the same answer. \citet{ko-etal-2020-look} found that BERT easily picks up the position bias in the SQuAD dataset \cite{rajpurkar-etal-2016-squad} and models' performance can drop by more than 50 points on \textit{f1}-score when training on a biased subset. \citet{sen-saffari-2020-models} analyzed the model's ability to generalize, by training on 5 different QA datasets, and found that no single dataset is robust to perturbations in the questions. \citet{shah-etal-2020-expect} tested models, trained on several multiple-choice QA datasets, and showed that they are largely relying on dataset biases. Previous work mostly studies the language models, as-is, or evaluated models fine-tuned on the QA datasets. In this paper, we go a step further and investigate the models adapted to a target task, using three different methods, and we study their effect on the model's learning process.

\begin{table*}
\centering
\caption{Results on the ProtoQA dev set, we ran experiments 5 times with different seeds, and report 95\% confidence interval. *Note that the human performance accuracies from \cite{boratko-etal-2020-protoqa} are reported on the test set, we assume that the accuracy values would be similar on dev set.}
\label{tab:results}
\resizebox{\linewidth}{!}{
\begin{tabular}{lccccccc}
\toprule
\textbf{Model} & \textbf{Ans@1} & \textbf{Ans@3} & \textbf{Ans@5} & \textbf{Ans@10} & \textbf{InCorr@1} & \textbf{InCorr@3} & \textbf{InCorr@5}\\
\hline
GPT2 & 28.2 & 27.1 & 27.2 & 30.7 & 14.4 & 21.1 & 27.5 \\
GPT2-Autoprompt & $25.5 (\pm 2.5)$ & $30.9 (\pm 2.8)$ & $35.1 (\pm 1.7)$ & $42.4 (\pm 1.7)$& $14.7 (\pm 2.6)$ & $28.7 (\pm 1.9)$ & $35.5 (\pm 1.3)$ \\ 
GPT2-Prefix-tuning & $42.7 (\pm 2.0)$  & $\bf 51.5 (\pm 1.4)$ & $52.2 (\pm 0.8)$ & $60.8 (\pm 0.5)$ & $28.8 (\pm 1.3)$ & $47.6 (\pm 0..9)$ & $56.9 (\pm 0.8)$ \\
GPT2-Finetune & $\bf 49.3(\pm 1.7)$  & $50.3 (\pm 1.5)$ & $\bf 53.0 (\pm 2.2)$ & $\bf 63.0 (\pm 0.6)$ & $\bf 31.9 (\pm 1.4)$ & $\bf 49.9 (\pm 1.4)$ & $\bf 57.9 (\pm 1.4)$ \\
\hline
BART & 20.9 & 29.8 & 32.2 & 37.5 & 15.1 & 27.3 & 32.2 \\
BART-Autoprompt & $28.2 (\pm 4.7)$ & $33.8 (\pm 0.9)$ & $37.2 (\pm 1.5)$ & $44.6 (\pm 3.0)$ & $16.6 (\pm 2.1)$ & $31.1 (\pm 2.3)$ & $38.9 (\pm 2.7)$ \\ 
BART-Prefix-tuning & $45.5 (\pm 2.9)$ & $51.0 (\pm 1.9)$ & $54.8 (\pm 1.6)$ & $\bf 62.9 (\pm 1.0)$ & $32.7 (\pm 1.4)$ & $51.4 (\pm 0.9)$ & $58.7 (\pm 1.0)$ \\
BART-Finetune & $\bf 53.6 (\pm 2.5)$ & $\bf 54.3 (\pm 2.2)$ & $\bf 56.3 (\pm 0.9)$ & $62.6 (\pm 1.0)$ & $\bf 35.6 (\pm 0.6)$ & $\bf 53.9 (\pm 1.5)$ & $\bf 59.5 (\pm 1.8)$ \\
\hline
Human* & 78.4 & 74.4 & 72.5 & 73.3 & 55.8 & 69.4 & 72.4 \\
Single Human* & 40.5 & 39.4 & 41.0 & 45.6 & 23.9 & 36.0 & 40.5 \\
\bottomrule
\end{tabular}
}
\end{table*}


\section{Experimental Setup}
\label{sec:experiments}
\subsection{Task and datasets}
\cut{\noindent\textbf{Task and datasets.}} We experiment with generative commonsense tasks, assuming that they are more realistic to real-world deployment of LMs and that they provide more insight about models' reasoning abilities. Specifically, we evaluate our models on the recently-introduced ProtoQA~\cite{boratko-etal-2020-protoqa} and CommonGen \cite{lin-etal-2020-commongen} datasets. For ProtoQA, given a question about a prototypical situation, the model is expected to produce a ranked list of answers. Each question in the dev and test sets is annotated with 100 answers, which are further manually grouped into clusters: the model's outputs are compared with the answer clusters, and the scores reflect the sizes of the matched clusters. We adopt ProtoQA's official evaluation metrics: \texttt{Max answer@$k$} (percentage of correct answers with top-$k$ predictions) and \texttt{Max Incorrect@$k$} (percentage of correct answers after making $k$ mistakes). We compute the answer matches based on WordNet similarity, as recommended by~\citet{boratko-etal-2020-protoqa}. For CommonGen, given a set of 3-5 input concepts, the task is to generate a scene description, utilizing all input concepts. Following prior work, we adopt \texttt{BLEU} \cite{10.3115/1073083.1073135}, \texttt{ROUGE} \cite{lin-2004-rouge}, \texttt{METEOR} \cite{banerjee-lavie-2005-meteor}, \texttt{CIDEr} \cite{VedantamZP15} and \texttt{SPICE} \cite{DBLP:journals/corr/AndersonFJG16}, as evaluation metrics.

\subsection{Strategies}
We describe how we adapt pre-trained GPT-2 and BART models to a target task with three methods\footnote{Our code is available at \url{https://github.com/Mayer123/CS_Model_Adaptation}}: \\
\textit{(S1) Fine-tuning} is the classic model adaptation approach, where all its parameters are updated using the training signal from the ground truth.\\
\textit{(S2) Prefix-tuning} \cite{li2021prefixtuning} is a method which fixes the pre-trained model's parameters during adaptation. This method adds trainable parameters, called prefix states, to the self-attention component~\cite{NIPS2017_3f5ee243} of every transformer layer in the model; only these prefix states are updated during training. Essentially, the prefix states act as conditioning variables that contextualize the representation of the inputs, such that the model can generate the desired outputs.\\
\textit{(S3)} Instead of updating model parameters, \textit{Autoprompt} \cite{autoprompt:emnlp20} appends a few trigger-tokens to the input and updates these trigger-tokens during training. Specifically, the gradient with respect to the trigger-tokens is computed using the ground-truth data. During training, new trigger-tokens are discovered, along the direction of the gradient, to replace the existing ones and to minimize the loss. Essentially, this method automatically learns to paraphrase the input question so that the model can generate the desired outputs.\\
We select fine-tuning, prefix-tuning, and Autoprompt as they are representative methods for adapting a pre-trained model to a target task, namely: 1. model adaptation (fine-tuning); 2. model extension (prefix-tuning); and 3. input adaptation (Autoprompt). We illustrate whether the model has learned different behaviors from methods with different degrees of adaptation. Training details can be found in the appendix \ref{sec:training}.

\subsection{Research Questions}
\cut{\noindent\textbf{Research questions.}} We address three questions in this paper, namely:\\ 
\textit{(RQ1: Adaptation level) How do different levels of adaptation affect the model's task-specific performance?} We expect that methods that adapt a larger number of parameters to the training task (fine-tuning) would perform better on the task itself, as the larger search space makes it more likely to find a task optimum. We investigate this by comparing S1-S3 on the two benchmarks. \\
\textit{(RQ2: Task structure) Do models only learn the task structure during training?} As we are working with relatively small benchmarks, we hypothesize that LMs acquire most of the necessary commonsense knowledge during pre-training instead of at adaptation time, during which they instead learn to \textit{elicit} this knowledge.
In this case, such adaptation to task structure could be done on just a subset of the training data without a large drop in performance, and the model need not depend on any lexical similarities between the training set and the dev set. To this end, we train our models with each adaptation method on: 1) a \textit{non-overlap} subset of ProtoQA, consisting of train-set QA pairs whose answers do not have any vocabulary overlap with the dev set answers; and 2) a 
\textit{min-overlap} split for CommonGen, selecting training instances whose input concepts appear at most once in the dev set. \\
\begin{table}
\caption{Summary of the dataset splits.}
\label{tab:data}
\centering
\begin{tabular}{ccc}
\toprule
\textbf{Dataset} & \textbf{Subset} & \textbf{\# of examples} \\
\hline
ProtoQA & Full-data  & 44,964 \\
ProtoQA & \textit{non-overlap}  & 18,855 \\
ProtoQA & \textit{similarity}  & 17,461 \\
\hline
CommonGen & Full-data  & 67,398 \\
CommonGen & \textit{min-overlap} & 17,914 \\
CommonGen & \textit{random}  & 17,914 \\
\bottomrule
\end{tabular}
\end{table} 
\textit{(RQ3: Novelty) Do models simply memorize the training data, or do they learn to reason on novel questions and answers as well?} 
To test whether models merely retrieve lexically similar examples, we formulate a \textit{similarity} subset on ProtoQA, comprised of the 100 questions in the training set with the highest cosine similarity for every dev set question. To test whether models achieve better performance due to improved \textit{reasoning} ability, we selected 30 questions from the ProtoQA dev set, where the model answers are at least partially correct, and we minimally changed the question through manual annotation---so that the required reasoning process is the same, but the answer set is different (example question pairs are given in the appendix \ref{sec:example}). Then, we use the BART model trained with each adaptation method to generate answers for the 30 new questions. We manually validate the 30 new questions and the 30 original questions, in order to compute the accuracy of the models as well as the percentage of overlapping answers between the original questions and new questions. More details are provided in section \ref{sec:manual}, and a summary of all the dataset splits used in our experiments is shown in Table \ref{tab:data}.

\begin{figure}[ht]
    \centering
    \includegraphics[width=\linewidth]{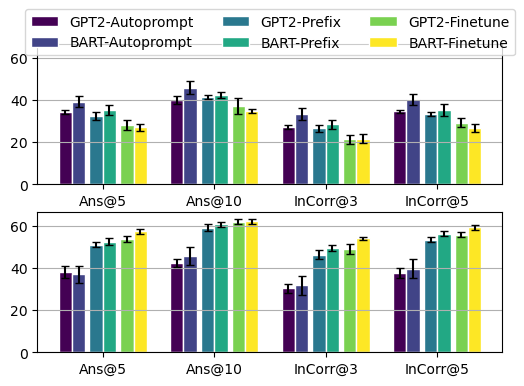}
    \caption{Results on ProtoQA dev set. The upper figure shows the models trained on the \textit{non-overlap} subset\cut{, which has 18,855 QA pairs}. The bottom figure shows the model trained on the \textit{similarity} subset\cut{, which has 17,461 QA pairs}.}
    \label{fig:robust}
\end{figure}


\section{Results}
\subsection{ProtoQA}
In response to \textit{RQ1 (adaptation level)}, Table \ref{tab:results} shows that prefix-tuning yields similar or slightly worse results compared to fine-tuning, for both LM classes, 
indicating that prefix-tuning is a promising lightweight alternative to fine-tuning. Autoprompt lags behind the tuning methods, while outperforming the zero-shot baseline. This is not surprising, as Autoprompt performs a fairly limited adaptation by only updating trigger tokens in a discrete space. 

\noindent The results for \textit{RQ2 (task structure)} are shown in Figure \ref{fig:robust}. Fine-tuning a model on the \textit{non-overlap} data leads to a drastic drop in performance, compared to using the full training data. Prefix-tuning's drop in performance is smaller than that of fine-tuning, while Autoprompt achieves the best performance when training on this subset. The result of Autoprompt is similar to training on the full data, showing that Autoprompt is much more robust towards an adversarial training split and is mainly learning how to elicit the model's pre-trained knowledge to answer the questions. 
Fine-tuning is learning knowledge together with the task structure, while prefix-tuning stands between fine-tuning and Autoprompt. Since prefix-tuning does not change the pre-trained model's parameters, but rather adds new parameters, it learns to mix the knowledge gained from pre-training with the signal from training instances to answer new questions. 
\noindent For \textit{RQ3 (novelty)}, the results of training on the \textit{similarity} subset are shown in Figure~\ref{fig:robust}. Although the number of QA pairs is much lower, fine-tuning achieves the same results as in the full-data setting. This shows that fine-tuning benefits more from the content of the training data than the task format, further informing our findings for \textit{RQ2}. Prefix-tuning performs slightly worse than the full data setting, indicating that here it is largely learning the training content. Autoprompt achieves similar results as in the full-data and \textit{non-overlap} settings, confirming our \textit{RQ2} observations that models are only learning the task format. We note that, while retrieving lexically similar questions might yield partial results, this form of pattern-matching is insufficient for commonsense reasoning.
For example, for a training question \textit{Name a vegetable that people like to steam}, the model learned the answer \textit{cauliflower}, which is coincidentally also a correct answer to the dev question \textit{Name a vegetable that is as large as your head}. In other words, the model answers correctly for the wrong reasons. \\

\begin{table}
\caption{Human evaluation results}
\label{tab:human}
\resizebox{\linewidth}{!}{
\centering
\begin{tabular}{lccc}
\toprule
\textbf{Measurement} & \textbf{Fine-tune} & \textbf{Prefix} & \textbf{Autoprompt} \\
\hline
Original question & 54.8  & 53.6  & 43.7 \\
New question & 42.3  & 42.3 & 30.9\\
\hline
\% overlap answers & 44.7   & 44.7  & 38.1 \\
\bottomrule
\end{tabular}}
\end{table}

\begin{table*}
\caption{Results on CommonGen dev set with the BART model.}
\label{tab:commongen}
\resizebox{\linewidth}{!}{
\centering
\begin{tabular}{lcccccc}
\toprule
\textbf{Method} & \textbf{Split} & \textbf{Bleu4} & \textbf{METEOR} & \textbf{ROUGE-L} & \textbf{CIDEr} & \textbf{SPICE}\\
\hline
Autoprompt & Full-data & $0.3 (\pm 0.2)$ &  $ 8.9 (\pm 3.0)$ &  $16.2 (\pm 4.1)$ &  $ 1.0 (\pm 0.5)$ & - \\
Prefix & Full-data & $34.4 (\pm 0.4)$ &  $\bf 33.0 (\pm 0.2)$ &  $53.9 (\pm 0.2)$ &  $\bf 17.4 (\pm 0.2)$ & $\bf 33.4 (\pm 0.3)$ \\
Fine-tune & Full-data & $\bf 34.9 (\pm 0.4)$ & $32.9 (\pm 0.1)$ & $\bf 54.1 (\pm 0.1)$ & $\bf 17.4 (\pm 0.1)$ &  $33.0 (\pm 0.2)$\\
\hline
Autoprompt & \textit{Min-overlap} & $0.3 (\pm 0.2)$ &  $ 8.0 (\pm 2.1)$ &  $16.5 (\pm 2.3)$ &  $ 1.1 (\pm 0.6)$ & - \\
Prefix & \textit{Min-overlap} & $\bf 32.8 (\pm 1.0)$ &  $\bf 31.5 (\pm 0.3)$ &  $\bf 52.1 (\pm 0.5)$ &  $\bf 16.6 (\pm 0.3)$ & $\bf 32.2 (\pm 0.2)$ \\
Fine-tune & \textit{Min-overlap} & $31.5 (\pm 0.2)$ &  $30.6 (\pm 0.3)$ &  $51.2 (\pm 0.2)$ &  $15.9 (\pm 0.1)$ & $30.7 (\pm 0.5)$ \\
\hline
Autoprompt & \textit{Random} & $0.1 (\pm 0.1)$ &  $ 7.4 (\pm 3.1)$ &  $14.5 (\pm 4.2)$ &  $ 1.1 (\pm 0.8)$ & - \\
Prefix & \textit{Random} & $33.4 (\pm 0.3)$ &  $\bf 32.3 (\pm 0.4)$ &  $52.9 (\pm 0.3)$ &  $\bf 17.2 (\pm 0.2)$ & $\bf 32.8 (\pm 0.7)$ \\
Fine-tune & \textit{Random} & $\bf 34.1 (\pm 0.3)$ &  $32.1 (\pm 0.1)$ &  $\bf 53.3 (\pm 0.2)$ &  $17.1 (\pm 0.2)$ & $32.3 (\pm 0.4)$ \\
\bottomrule
\end{tabular}}
\end{table*}

\subsubsection{Manual Annotation} 
\label{sec:manual}
We conduct manual annotation, to further verify our observations for \textit{RQ3}. For the BART model, trained with each adaptation method, we generate top10 answers for every question and then annotate each answer, independently. We annotate each answer on a 5-point Likert scale, where 1 means strongly disagree, 2 means mostly disagree, 3 means not sure/it depends, 4 means mostly agree, and 5 means strongly agree.
In total, 4 researchers annotated 1165 QA pairs where each QA pair received 3 ratings. The overall Kripendorf alpha \cite{krippendorff04} score is 0.52, which is moderate agreement. If we merge answers choices 1 and 2 to be `incorrect' and merge 4 and 5 to be `correct', and then compute the 3-class categorical agreement score using Fleiss kappa, the agreement score is 0.36, which is fair agreement. 
Then we consolidate the 3 ratings by taking the average of the 3 annotations and consider an answer to be correct if the average score is greater than 3.5.

The results from manual assessment of the models' reasoning capabilities are shown in Table \ref{tab:human}. We observe that our LMs are not able to capture subtle changes in the question that lead to a different answer set; models are getting worse performance on the new questions, overall. We believe this is because the newly-generated questions are more difficult to answer, as they seldom appear in any text corpus in general. 
We also see a high overlap between the generated answers to the original and the newly-created questions, especially for fine-tuning and prefix-tuning, where nearly half (44.7\%) of the answers are repeated. This confirms our observation that models memorize/retrieve training-set answers without actually engaging in reasoning. 


\subsection{CommonGen}
\label{sec:commongen}
The full results on the CommonGen dataset are shown in Table \ref{tab:commongen}. Overall, we can see that the results follow a similar trend to those of ProtoQA, as prefix-tuning is able to perform significantly better than fine-tuning when trained on an `adversarial' split. We notice that the relative drop of performance for both methods on the \textit{Min-overlap} subset is less drastic than that of ProtoQA. We think this is mainly due to the task format. For ProtoQA, models need to perform one or a few hops of reasoning to answer the questions and there is no direct evidence from the question itself, i.e., the model cannot directly copy answers from the questions. However, the model is directly given the target concepts as inputs for CommonGen, which the model can directly use as its outputs. Thus, we argue that the amount of reasoning required in CommonGen is more restricted than in ProtoQA, and models are less likely to leverage the clues to solve the task. \\
Also, it is worth noting that the accuracy of Autoprompt is extremely low on all 3 splits. In fact, Autopropmt fails to generate any meaningful sentences after training, and the \texttt{SPICE} metric could not be computed. We, again, attribute this to the task format. Autoprompt would eventually discover tokens that are meaningless to humans, and we can think of them as injecting task-specific noise to the pre-trained models. For ProtoQA, the model is expected to generate single words or short-phrase answers to complete the sentence, i.e., converted question, thus it is reasonable for the model to do it even with the injected noise. However, for CommonGen, the model is expected to generate a full sentence as output; with Autoprompt, the task basically translates to generating a sentence given input concepts and a bunch of random tokens, which is very different from BART's pre-training context.


\section{Conclusions}
Experiments with two language model classes, on two generative commonsense benchmarks, under three adaptation methods, revealed that the learning efficiency of LMs relies heavily on the adaptation method. Fine-tuning teaches the model both the structure of the task and the content, prompting approaches focus on learning the task structure only, while model extension by prefix-tuning falls between these two extremes. Consequently, prompting is the least sensitive of the three methods to the training data size and quality, and prefix-tuning can generalize better to novel concepts regardless of the task format. Future work on generalizable common sense leverage these findings, and: 1) avoid fine-tuning, as we may never be able to create datasets without any unintended biases~\cite{linzen-2020-accelerate}; and 2) evaluate on multiple independent test-sets to better replicate real-world settings, as training on any split of data can lead to an overestimation of performance~\cite{sogaard-etal-2021-need}. 


\bibliography{anthology}
\bibliographystyle{acl_natbib}
\clearpage

\appendix
\section{Appendix}
\label{sec:appendix}

\subsection{Training details}
\label{sec:training}
\subsubsection{Hyper-parameters}
For all of our experiments on ProtoQA, we follow the baseline model setup of~\citet{boratko-etal-2020-protoqa} and train the model with learning rate 1e-5, batch size 8, warm-up steps 150 and Adam epsilon 1e-6, unless otherwise specified. We trained the model for the amount of steps that are equivalent to 1 epoch of training on full training set, i.e., we train the model for longer epochs on \textit{non-overlap} and \textit{similarity} subsets.
For both training and inference, we adopted the same question-converting templates as \citet{boratko-etal-2020-protoqa}. During inference, we do nucleus sampling with top-p=0.9, temperature 0.69 to generate 300 answer candidates; we group them and rank them based on frequency. \\
For all of the experiments on CommonGen, we used learning rate of 1e-5, batch size 16, warm-up steps 500 and Adam epsilon 1e-6. We trained the model for 2 epochs and similarly we train models for longer epochs on \textit{min-overlap} and \textit{random} subsets. During inference, we do beam search with beam size 5, length penalty 0.6, and repetition penalty 2.0. Note that we disabled positional embeddings in the BART encoder, for all CommonGen experiments, as we found them detrimental to model performance.  \\
\subsubsection{Model Implementation}
We used the BART-large and GPT2-large model provided by the transformers library \cite{Wolf2019HuggingFacesTS}.
For prefix-tuning, we used prefix with length 10 and a 1 layer of prefix MLP with hidden size 512 (we tried \{512, 800\} and found them to have very close results). The learning rate is 5e-5, while other hyperparameters are the same as in fine-tuning (we tried \{1e-5, 2e-5, 5-e5, 8e-5\}, and found the latter 2 achieve slightly better results). For prefix-tuning with the BART model, we added prefix states to self-attention in encoder layers and self-attention and cross-attention in decoder layers. For GPT2 model, we only add prefix states to self-attention in decoder layers. \\
For Auto-prompt with BART, we used the same 10 trigger tokens for both encoder and decoder; the trigger tokens are all initialized with mask tokens. For GPT2, we also used 10 trigger tokens. Since the model does not have mask tokens, we initialized triggers with the tokenized prompt "Based on simple commonsense fact, we know that", which is exactly 10 tokens by BPE. We train the models with batch size 32 and gradient accumulation steps 4 (we tried batch sizes \{32, 128, 256\} and found that larger batch size yield more stable results). At each update step, we search for the next trigger token, within the 100 closest candidate tokens, along the gradient direction (we used 10 candidate tokens for CommonGen experiments, as we found that both 10 and 100 lead to extremely bad results---so we used 10 to save computation time). \\A summary of the number of trainable parameters for each model-adaptation method combination is shown in Table \ref{tab:params}. 

\begin{table}
\caption{Trainable parameters comparison}
\label{tab:params}
\centering
\begin{tabular}{ccc}
\toprule
\textbf{LM} & \textbf{Adaptation} & \textbf{\# Parameters} \\
\hline
GPT2 & Fine-tune  & 774M \\
GPT2 & Prefix-tuning  & 900K \\
GPT2 & Autoprompt  & 0 \\
\hline
BART & Fine-tune  & 400M \\
BART & Prefix-tuning  & 1.44M \\
BART & Autoprompt  & 0 \\
\bottomrule
\end{tabular}
\end{table}

\subsubsection{Dataset splits}

The ProtoQA dataset provides a dev-scraped set and a dev-crowdsourced set, where dev-scraped is collected from the Family-feud fan website, i.e., same as the training set, while the dev-crowdsourced set contains newly written questions and answers by crowd-workers, i.e., same as the test set. We select the best model using the loss on dev-scraped set and report results on the dev-crowdsourced set, because the test set answers are hidden and we need the ground-truth answers to test our hypothesis. In the main paper content, all references of ProtoQA dev-set refer to the dev-crowdsourced set.
For CommonGen dataset, we select best models, using the loss on the dev-set. 

For the \textit{similarity} subset of ProtoQA, we adopt the \texttt{stsb-roberta-large} model from the sentence-transformer \cite{reimers-2019-sentence-bert} library and compute the cosine similarity between the train and the dev questions.

\subsection{New Questions}
\label{sec:example}
Examples of the original and newly written questions, along with the model predictions are shown in Table \ref{tab:example}. 

\begin{table*}
\caption{Example of original questions and newly written questions and the corresponding predictions from BART model trained on Full data, the bold answers are corrected ones selected by human evaluation}
\label{tab:example}
\resizebox{\linewidth}{!}{
\centering
\begin{tabular}{ll}
\toprule
\textbf{Type} & \textbf{Question/Answers} \\
\hline
Original  & \bf Name something a monk probably would not own. \\
Fine-tune & \textbf{car}, clothes, money, bike, \textbf{computer}, horse, \textbf{cell phone}, shoes, \textbf{motorcycle}, bicycle \\
Prefix-tuning & \textbf{car}, \textbf{motorcycle}, bike, clothes, money, robe, shoes, \textbf{sword}, horse, \textbf{camera} \\
Autoprompt & money, \textbf{car}, \textbf{gun}, \textbf{sword}, \textbf{boat}, wine, shoes, coffee, \textbf{beer}, cat \\
\hline
New & \bf Name something a monk probably would own. \\
Fine-tune & \textbf{pen}, cross, \textbf{book}, sword, \textbf{rosary}, brooms, bible, \textbf{robe}, \textbf{books}, kite \\
Prefix-tuning & \textbf{robe}, bible, cross, \textbf{pen}, \textbf{beard}, rosa, sword, \textbf{robes}, \textbf{monastery}, apron \\
Autoprompt & money, bike, sword, \textbf{chair}, \textbf{books}, \textbf{brooch}, bicycle, gold, phone, kitty \\
\hline
\hline
Original  & \bf Name a vegetable that is about as big as your head.. \\
Fine-tune & broccoli, carrot, \textbf{cabbage}, \textbf{cauliflower}, carrots, \textbf{lettuce}, spinach, \textbf{cucumber}, potato, corn \\
Prefix-tuning & \textbf{cucumber}, broccoli, \textbf{cabbage}, carrot, \textbf{lettuce}, \textbf{pumpkin}, spinach, beet, \textbf{cauliflower}, celery \\
Autoprompt & \textbf{cabbage}, broccoli, spinach, \textbf{lettuce}, carrots, \textbf{cauliflower}, potato, potatoes, \textbf{pumpkin}, gooseberries \\
\hline
New & \bf Name a vegetable that is about as big as your fist. \\
Fine-tune & broccoli, cabbage, carrot, carrots, spinach, lettuce, cauliflower, cucumber, corn, \textbf{beet} \\
Prefix-tuning & cucumber, broccoli, cabbage, lettuce, spinach, \textbf{beet}, celery, pumpkin, carrot, cauliflower \\
Autoprompt & broccoli, cabbage, lettuce, carrots, spinach, squash, cucumbers, eggplant, cucumber, \textbf{kiwi} \\
\hline
\hline
Original  & \bf Name a sport that requires a lot of equipment. \\
Fine-tune & \textbf{hockey}, \textbf{golf}, \textbf{football}, \textbf{tennis}, basketball, \textbf{baseball}, soccer, boxing, wrestling, volleyball \\
Prefix-tuning & \textbf{football}, basketball, \textbf{hockey}, soccer, \textbf{tennis}, \textbf{golf}, \textbf{baseball}, wrestling, volleyball, \textbf{skiing} \\
Autoprompt & \textbf{hockey}, soccer, \textbf{golf}, basketball, \textbf{football}, \textbf{tennis}, baseball, \textbf{rugby}, volleyball, \textbf{ice hockey} \\
\hline
New & \bf Name a sport that you don't need a lot of equipment for. \\
Fine-tune & hockey, tennis, baseball, golf, \textbf{soccer}, football, \textbf{basketball}, \textbf{bowling}, \textbf{volleyball}, \textbf{swimming} \\
Prefix-tuning & \textbf{basketball}, tennis, \textbf{soccer}, golf, football, hockey, baseball, \textbf{bowling}, \textbf{swimming}, skiing \\
Autoprompt & \textbf{basketball}, \textbf{soccer}, hockey, golf, \textbf{volleyball}, tennis, football, baseball, rugby, lacrosse \\
\hline
\hline
Original  & \bf Name something around the house that's often replaced. \\
Fine-tune & tv, television, furniture, \textbf{dishes}, carpet, \textbf{toilet paper}, refrigerator, windows, stereo, \textbf{lights} \\
Prefix-tuning & carpet, lamp, light, furniture, tv, clothes, \textbf{dishes}, television, \textbf{bedding}, \textbf{toilet paper} \\
Autoprompt & TV, tv, couch, table, toilet, bed, television, microwave, chair, lamp \\
\hline
New & \bf Name something around the house that's hardly ever replaced. \\
Fine-tune & tv, television, \textbf{furniture}, dishes, \textbf{refrigerator}, stereo, \textbf{carpet}, toilet paper, \textbf{windows}, \textbf{appliances} \\
Prefix-tuning & dishes, \textbf{furniture}, lamp, \textbf{carpet}, tv, clothes, bedding, TV, light, television \\
Autoprompt & TV, tv, \textbf{fridge}, \textbf{microwave}, \textbf{couch}, \textbf{refrigerator}, \textbf{dishwasher}, coffee table, \textbf{bed}, \textbf{table} \\
\hline
\hline
Original  & \bf Name a job where you have to be awake at night. \\
Fine-tune & \textbf{police officer}, doctor, \textbf{nurse}, \textbf{security guard}, lawyer, teacher, \textbf{firefighter}, construction, actor, \textbf{cop} \\
Prefix-tuning & \textbf{police officer}, \textbf{nurse}, construction, doctor, \textbf{security guard}, \textbf{bartender}, waiter, babysitter, \textbf{firefighter}, teacher \\
Autoprompt & construction, work, carpenter, \textbf{firefighter}, \textbf{truck driver}, roofing, \textbf{police}, \textbf{fireman}, \textbf{bartender}, school \\
\hline
New & \bf Name a job where you only have to work during the day. \\
Fine-tune & nurse, \textbf{teacher}, police officer, \textbf{doctor}, bartender, construction, waitress, \textbf{lawyer}, waiter, \textbf{mechanic} \\
Prefix-tuning & nurse, \textbf{teacher}, \textbf{lawyer}, \textbf{doctor}, bartender, waiter, \textbf{mechanic}, construction, \textbf{sales}, waitress \\
Autoprompt & construction, hospital, fireman, restaurant, \textbf{plumber}, cook, \textbf{cleaning}, chef, firefighter, \textbf{teaching} \\
\hline
\bottomrule
\end{tabular}}
\end{table*}

\end{document}